\documentclass{article}

\usepackage[preprint]{neurips_2025}

\usepackage{amsmath} 
\usepackage{adjustbox}
\usepackage{multirow}
\usepackage{graphicx}
\usepackage{wrapfig}
\usepackage{booktabs}

\usepackage{float} % Required for the 'H' placement option
\usepackage[utf8]{inputenc} % allow utf-8 input
\usepackage[T1]{fontenc}    % use 8-bit T1 fonts
\usepackage{hyperref}       % hyperlinks
\usepackage{url}            % simple URL typesetting
\usepackage{booktabs}       % professional-quality tables
\usepackage{amsfonts}       % blackboard math symbols
\usepackage{nicefrac}       % compact symbols for 1/2, etc.
\usepackage{microtype}      % microtypography
\usepackage{xcolor}         % colors
\usepackage{enumitem}

\title{Plan-and-Refine: Diverse and Comprehensive Retrieval-Augmented Generation}

\author{Alireza Salemi, Chris Samarinas, Hamed Zamani \\
  University of Massachusetts Amherst \\
  \texttt{\{asalemi, csamarinas, zamani\}cs.umass.edu}}

\usepackage{xspace}

\newcommand{\ourmethod}{\texttt{P\&R}\xspace}

\begin{document}

\setcitestyle{numbers, square}
\maketitle

\begin{abstract}
This paper studies the limitations of (retrieval-augmented) large language models (LLMs) in generating diverse and comprehensive responses, and introduces the Plan-and-Refine (\ourmethod) framework based on a two phase system design. In the global exploration phase, \ourmethod generates a diverse set of plans for the given input, where each plan consists of a list of diverse query aspects with corresponding additional descriptions. This phase is followed by a local exploitation phase that generates a response proposal for the input query conditioned on each plan and iteratively refines the proposal for improving the proposal quality. Finally, a reward model is employed to select the proposal with the highest factuality and coverage. We conduct our experiments based on the ICAT evaluation methodology--a recent approach for answer factuality and comprehensiveness evaluation. Experiments on the two diverse information seeking benchmarks adopted from non-factoid question answering and TREC search result diversification tasks demonstrate that \ourmethod significantly outperforms baselines, achieving up to a 13.1\% improvement on the ANTIQUE dataset and a 15.41\% improvement on the TREC dataset. Furthermore, a smaller scale user study confirms the substantial efficacy of the \ourmethod framework.
\end{abstract}

\section{Introduction}
\label{sec:introduction}

LLMs have shown strong performance in text generation by producing fluent, coherent, engaging, and contextually relevant responses to their prompts \citep{eldan2023tinystoriessmalllanguagemodels, wang-etal-2023-document-level, gomez-rodriguez-williams-2023-confederacy, HadiLargeLM}. To address the well-known hallucination issue and deal with non-stationary and up-to-date information, state-of-the-art question answering systems as well as generative models enhance LLMs through retrieval augmentation \citep{reml}, an approach commonly referred to as retrieval-augmented generation (RAG). 
However, recent studies reveal that the text generated by RAG models still generate non-factual content \citep{min-etal-2023-factscore}, and more importantly, they lack response diversity and comprehensiveness \citep{icat}. This is while ensuring accurate, diverse, and comprehensive responses is essential for applications such as non-factoid question answering, exploratory search, information seeking in domains such as healthcare, legal assistance, education and research, and information-driven decision making.

% This is while ensuring that responses generated by LLMs are complete and accurate is essential to mitigating the spread of misinformation, achieving diverse and thorough coverage of all important aspects of a given input, and fostering user trust in the system \citep{icat,gao-etal-2023-enabling}. 

This paper addresses this gap by proposing methods that satisfy two key desiderata: (1) diversity and comprehensiveness — model outputs should capture and respond to the full range of relevant aspects of the input question, and (2) factuality — the responses must consist of factually accurate claims. While the concept of novelty and diversity in retrieval has been extensively explored within the information retrieval community \citep{mmr, 10.5555/646180.682439, Leelanupab2012ARF, 10.1145/2063576.2063869}, training LLMs and RAG systems to generate diverse and comprehensive responses to their input query is relatively underexplored. 

Recent work by \citet{icat} shows that state-of-the-art LLMs not only sometimes generate non-factual content, but also do not perform well in generating comprehensive responses, even if they are specifically asked to in their prompts. This observation has also been validated in our experiments. We also observe that diversifying the retrieval results in the RAG pipelines does not improve response coverage. This deficiency in generating comprehensive and factual responses arises from several factors. First, the pre-training sequence-to-sequence objectives \citep{seq2seq} and post-training techniques \citep{rlhf, rest, rest-em} are not specifically designed to encourage the generation of diverse outputs. We observe that techniques like Chain-of-Thought (CoT) prompting \citep{cot, liu2023logicot, 10.1007/978-981-97-7232-2_13} that perform well in mathematical reasoning tasks, fall short in improving response diversity and completeness. Second, the prevalent autoregressive generation paradigm, which relies on greedy decoding or sampling-based token selection, is inherently limited. It tends to favor locally optimal token predictions, often overlooking factual and comprehensive completions that diverge from the initial token prefix. This token-by-token generation process exacerbates the influence of early poor token choices, potentially distorting the response structure and leaving critical elements inadequately addressed.

\begin{wrapfigure}{r}{0.55\textwidth}
    \centering
    % \hspace{-0.4cm}
    \vspace{-0.4cm}
    \includegraphics[width=0.55\textwidth]{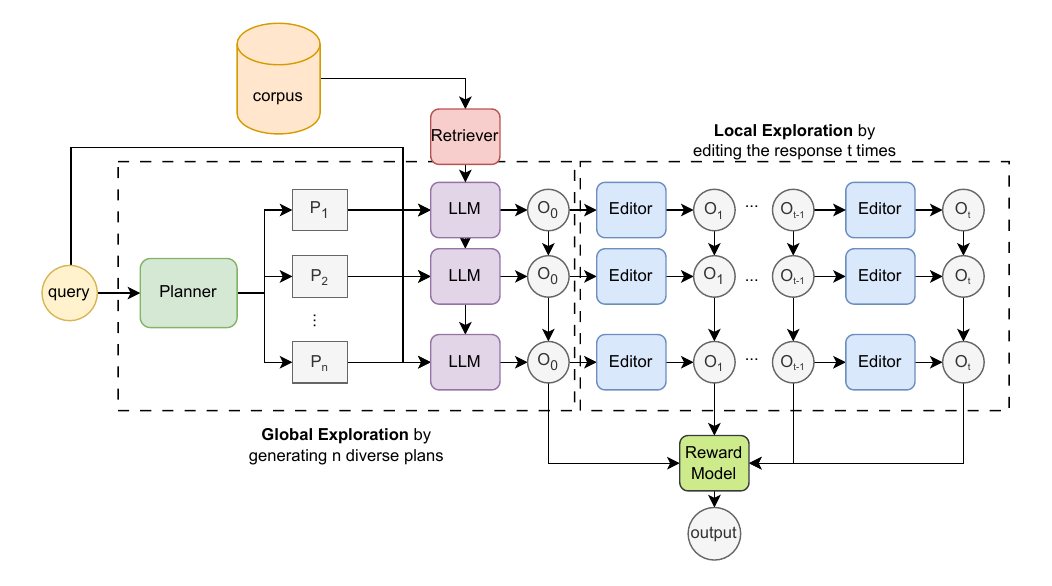}
    \vspace{-0.8cm}
    \caption{An overview of the \ourmethod framework.}
    \vspace{-0.2cm}
    \label{fig:overview}
\end{wrapfigure}

This paper introduces the \textit{Plan-and-Refine (\ourmethod)} framework to address both of these issues. \ourmethod is generic and can be applied to any RAG pipelines.\footnote{\ourmethod is model-agnostic; We focus on RAG because retrieval is key to generating factual, up-to-date responses.} An overview of this framework is presented in Figure~\ref{fig:overview}. \ourmethod consists of two main phases. We refer to the first phase as \textit{``planning''}, which generates a set of diverse plans for \textit{global exploration}. Each plan includes a list of diverse query aspects essential for creating a comprehensive response, the reasoning behind the value and relevance of each aspect, and the corresponding query of each aspect for retrieving diverse and relevant information about the aspect. A diverse set of plans are created by a \textit{planner} that we optimize via self-training, enabling it to identify diverse key aspects and later structure comprehensive responses effectively. Using each generated plan and the retrieved information for the plan, an LLM generates a detailed response to the query. Therefore, each plan results in a potential response for the query. This global exploration phase is followed by a refining phase as \textit{local exploitation}. This second phase, refine the LLM response multiple times to improve its comprehensiveness and factuality conditioned on the given plan. Finally, \ourmethod uses a trained reward model to evaluate all generated refinements, selecting the one with the highest factuality and coverage. This ensures the final output is the most accurate and comprehensive response to the input query.

% To overcome the limitations of greedy sampling mentioned above, \ourmethod introduces a multi-step process to enhance response diversity, factuality, and comprehensiveness. First, \ourmethod samples a diverse set of plans from the planner, each representing a potential solution to the query. This phase serves as a global exploration, examining a broad range of possible solutions (i.e., plans) to respond the query comprehensively. Next, to refine the initial responses, \ourmethod employs a self-trained editing model. Given a plan and its associated response, the editing model iteratively improves the response by enhancing its depth, factual accuracy, and coherence. This phase acts as a local exploration within the response space defined by each plan, improving the quality of responses. Finally, \ourmethod utilizes a trained reward model to evaluate all the generated responses. The reward model selects the response with the highest factuality and coverage, ensuring the final output provides the most accurate and comprehensive answer to the query.

We conduct our experiments on two diverse information-seeking tasks that benefit from comprehensive responses. We use ANTIQUE \citep{antique}---the largest non-factoid question answering dataset with complete manual relevance judgments---and the TREC Web Track data from 2009 to 2012 \citep{trec09, trec10, trec11, trec12} which is based on ClueWeb09 English documents. TREC Web Track ran a successful search result diversification task during this period, meaning that the queries in the dataset have multiple aspects and benefit from diverse perspectives. We used the ICAT evaluation methodology \citep{icat} for evaluating factuality and information coverage in the generated text in respnse to queries in these datasets.
Our results show that \ourmethod outperforms a competitive and diverse set of open-source and proprietary baselines across both datasets, achieving a statistically significant relative improvement of 13.1\% on the ANTIQUE dataset and 15.4\% on the TREC datasets. We further conducted a small user study to demonstrate user's preferences over the best baseline model. We observe that in 63\% of cases, annotators prefer \ourmethod's responses over the ones produced by our best performing RAG baseline with the same LLM.  To foster research in this area, we release our codebase.\footnote{Available at: \url{https://github.com/alirezasalemi7/PR-RAG}}

\section{Related Work}

\paragraph{\textbf{Retrieval-Augmented Generation.}}

RAG \citep{rag, urag, fid} integrates retrieval and generation to enhance the quality and relevance of content by using external knowledge during generation \citep{asai2023selfrag, siriwardhana-etal-2023-improving, salemi2024learningrankmultipleretrievalaugmented}. Unlike traditional LLMs, which rely solely on pre-trained knowledge, RAG systems retrieve information from a corpus via a retriever, enabling them to produce more contextually accurate outputs \citep{reml, kim2024retrievalenhancedmachinelearningsynthesis, salemi2024evaluating}. RAG's versatility allows it to be applied across various domains, such as knowledge grounding in textual \citep{kilt, rag, fid, StochasticRAG} and multi-modal \citep{dedr, murag, kat, salemi2023pretraining}, personalization \citep{salemi2024lamp,xu2025personalizedgenerationlargemodel, salemi2024optimization, salemi2024comparingretrievalaugmentationparameterefficientfinetuning, salemi2025reasoningenhancedselftraininglongformpersonalized, kumar2024longlampbenchmarkpersonalizedlongform, salemi2025experteffectiveexplainableevaluation}, and reducing hallucination \citep{agrawal2023knowledge, shuster-etal-2021-retrieval-augmentation}. We use RAG to improve factual coverage of generated responses.

\paragraph{\textbf{Planning \& Reasoning in Text Generation.}}

Solving complex problems often involves breaking them into subproblems, each tackled independently \citep{hu2024agentgenenhancingplanningabilities, planning-code, huang2024understandingplanningllmagents}. This is seen as planning a sequence of simpler steps to address a larger challenge. Subproblems frequently require reasoning---step-by-step processing referred to as chain-of-thought (CoT). CoT has been shown to improve LLMs on tasks involving mathematical, logical, and commonsense reasoning \citep{cot, liu2023logicot, 10.1007/978-981-97-7232-2_13}, and has been applied in evaluation \citep{kaneko2024evaluatinggenderbiaslarge}, code generation \citep{code-reasoning}, alignment \citep{reasoning-aligning}, and personalization \citep{salemi2025reasoningenhancedselftraininglongformpersonalized}. While reasoning in free-form text generation remains underexplored \citep{reasoning-survey}, recent work highlights its potential for generating high-quality text in emotionally and personally nuanced contexts \citep{salemi2025reasoningenhancedselftraininglongformpersonalized, li2024enhancingemotionalgenerationcapability}. We employ planning with reasoning to enhance the comprehensiveness and factuality of generated responses.

%Specifically, we leverage planning to explore the global space of complete and factual solutions to the input prompt prior to generating the final response. Following this, reasoning is applied to assess and execute each step in the solution, ensuring factuality and comprehensiveness of responses.

\paragraph{\textbf{Diversity \& Coverage.}}

Diversity have been extensively studied in the retrieval community \citep{8621577, Calumby2016DiversityorientedMA, 10.5555/646180.682439}, with several TREC tracks dedicated to it \citep{trec09, trec10, trec11, trec12}. Traditionally, these concepts have been approached as syntactical problems in text generation, where diversity is often evaluated based on the variety of words and phrases using n-gram metrics \citep{diversity-metric-ngram}, with less attention given to content diversity \citep{tevet-berant-2021-evaluating}. Consequently, much of the prior work has focused on improving syntactical diversity \citep{diverse-gen-penealize-reapeated, shao-etal-2019-long}.
% Recently, the TREC RAG track has also introduced the concept of evaluating text generation coverage through Nuggets \citep{pradeep2024initialnuggetevaluationresults}. However, judgments on this aspect are not yet publicly available. 
In parallel, ICAT \citep{icat} has been introduced as a metric that evaluates diversity, completeness, and factuality of generated responses based on their content rather than syntax. This paper focuses on enhancing the diversity, coverage, and factuality of LLM-generated outputs in terms of content.

\paragraph{\textbf{Scaling Test-Time Compute.}}

Recent advances in LLM reasoning for logical and mathematical tasks have shown that increasing the compute budget during inference enhances performance \citep{snell2024scalingllmtesttimecompute, chen2024simpleprovablescalinglaw}. This allows LLMs to utilize additional inference resources to explore the response space, providing more accurate answers in tasks such as code generation, logical, and mathematical reasoning \citep{bi2024forestofthoughtscalingtesttimecompute, zhang2024scalingllminferenceoptimized, brown2024largelanguagemonkeysscaling}. Most prior research has focused on math, code, and logic, with limited exploration in free-form text generation. We extend this concept to free-form generation by using enhanced inference compute to better search the response space and produce more comprehensive, factual responses.

\section{Problem Formulation}

A generative language model \( M_G \) takes a prompt \( x \)  and produces \( \bar{y} \) as the response. The quality of the generated output can be assessed based on various factors; coherence, accuracy, relevance, fluency, and alignment. One aspect that has received relatively little attention is \textit{comprehensiveness} while maintaining \textit{factuality} \citep{icat}. In this context, the response should offer a comprehensive and thorough coverage of topics related to the input, ensuring the output remains factually accurate and minimizes incorrect information based on a reference knowledge corpus \( C \). This corpus can take various forms, such as unstructured text, an encyclopedia, or even the entire web, as long as it comes from a trusted source. The main goal of this paper is to improve LLMs' ability to generate responses that are both highly factual and comprehensive. We assume that this quality can be quantified using a utility function or evaluation metric \( \mu \). Specifically, we employ ICAT \citep{icat} as the evaluation metric to assess the coverage of diverse factual information in long-form text generation. We aim to enhance LLMs' ability to achieve higher ICAT scores, thereby improving the quality of their output in terms of factual coverage. We assume access to a set of training queries that benefit from comprehensive and diverse responses \( D_{\text{train}} = \{x_i\}_{i=1}^{|D_{\text{train}}|} \) and a set of validation queries \( D_{\text{test}} = \{x_i\}_{i=1}^{|D_{\text{test}}|} \), both consist solely of input prompts without any corresponding reference outputs. Such queries can be obtained from non-factoid question answering datasets, community question answering websites, and discussion forums. In this setup, we propose methods to improve \( \mu \) in a reference-free setting.

\section{The \ourmethod Framework}

Ensuring accurate and complete LLM responses is vital to prevent misinformation and build user trust \citep{icat,gao-etal-2023-enabling}. As discussed in Section~\ref{sec:introduction}, prior research shows that LLMs struggle to consistently produce complete and accurate responses. Even very capable models like GPT-4 \citep{gpt4} cover less than 50\% of relevant subtopics on average for a given prompt \citep{icat}. Furthermore, while RAG enhances factuality, we show it reduces the coverage of generated responses (see Section~\ref{sec:main-findings}). Several factors contribute to this. Current pre-training  \citep{seq2seq} and post-training objectives \citep{rlhf, rest, rest-em} do not effectively encourage factual and comprehensive responses. Even techniques like CoT prompting \citep{cot,liu2023logicot, 10.1007/978-981-97-7232-2_13}, designed for mathematical reasoning, fail to improve response completeness. Moreover, the token-by-token text generation approach is sub-optimal. It can often overlook factual and complete responses that deviate from the prefix of generated tokens, leading to incomplete outputs. In essence, the LLM's initial token selection influences output structure, causing key aspects to be missed or underrepresented.

A simple solution to these issues could be to explicitly instruct LLMs to generate complete and factual responses that consider all aspects of the question. Furthermore, post-training techniques such as RLHF \citep{rlhf} or Self-Training \citep{rest, rest-em} can be used to optimize a reward model that accounts for the completeness of the response. However, they do not address the inherent problem of sampling responses from LLMs, where the structure of the output can still be influenced by the initial tokens, potentially leading to incomplete or inaccurate responses. To address the aforementioned challenges, we introduce \ourmethod, a novel approach that first generates a set of plans outlining the aspects that need to be covered, along with the rationale about why each aspect is important for a complete and factual response to the prompt and the query to retrieve information about each aspect. The model then generates responses based on each plan and retrieved documents and iteratively refines them through multiple editing steps. Finally, a reward model is employed to select the response with the highest score as the final output. 
The following subsections offer a detailed explanation of this approach.

\paragraph{\textbf{Overview.}} 

The overview of \ourmethod is shown in Figure~\ref{fig:overview}. We assume the existence of a planner \( M_P(x) \) that takes the input prompt \( x \) and returns a plan $p$ for generating factual and complete responses to the prompt. The plan \( p = \{(a_i, q_i, r_i)\}_{i=1}^{|p|} \) consists of a set of aspects \( a_i \) about the prompt, a query \( q_i \) to gather information about the respective aspect, and a reason \( r_i \) explaining why this aspect is important for a complete and factual response to the prompt. We assume the existence of a retriever \( R \) and a retrieval budget \( k \) to collect the necessary information for improving the factuality of the claims in the response. To gather the necessary information for the plan \( p \), for each \( (a_i, q_i, r_i) \in p \), we retrieve \( \frac{k}{|p|} \) documents for the query \( q_i \) from the corpus \( C \). They together form the context \( I_p = \bigcup_{(a_i, q_i, r_i)\in p} R(q_i, \frac{k}{|p|}, C) \) that can be used during response generation. We assume the existence of a generative model \( M_G(x, p, I_p) \) that, given a prompt \( x \), a plan \( p \), and the context \( I_p \), generates an output response \( o_p \) to the prompt with the given plan $p$ as the steps to take. To explore diverse solutions to the problem, we sample \( n \) distinct plans using the planner \( M_P \), resulting in a set of plans \( P = \{p_i\}_{i=1}^{n} \). These plans provide a range of strategies that to address the problem effectively. This step can be seen as sampling and searching through the space of all potential solutions to the problem, a process we refer to as \textbf{Global Exploration}. Then, the generative model \( M_G \) is applied to each plan \( p \in P \), producing an initial set of proposed responses \( O_0 = \{M_G(x, p, I_p) \mid p \in P\} \). 

While global exploration generates a diverse set of solutions to the prompt, it often falls short in meeting specific requirements with precision. To address this, we introduce the concept of \textbf{Local Exploitation}, which focuses on refining these solutions through targeted adjustments. This approach enhances and ensures higher-quality responses. For this, we assume the existence of an editing model \( M_E \) that, given the input prompt \( x \), a plan \( p \), and a previously generated response \( o_{t-1} \) for this prompt and plan, improves the response to generate \( o_t = M_E(x, p, o_{t-1}) \). Using this iterative approach, we can refine the initial set of generated responses. At each step, the updated responses are represented as \( O_t = \{M_E(x, p, o_{t-1}) \mid o_{t-1} \in O_{t-1}, p \in P\} \). By repeating this editing process \( T \) times, we obtain a final set of response proposals, denoted as \( O_F = \bigcup_{t=0}^{T} O_t \), which encompasses all the initial set of responses and refined outputs generated in iterations. Finally, to identify the most suitable response among all proposed candidates, we need to employ a mechanism to select the one that best meets the prompt's requirements, prioritizing completeness and factuality---key objectives of this problem. We assume the existence of a reward model \( M_R(x, o) \) that assigns a score to each generated output \( o \in O_F \) based on the input prompt \( x \). The final response to the prompt is selected as the output that achieves the highest score according to the reward model, formally as:  $o_f = {\mathrm{arg\,max}}_{{o \in O_F}} \, M_R(x, o)$. This ensures the chosen response is the most complete and factual among the generated candidates.

\subsection{Global Exploration through Planning}

We define a plan for responding to a prompt \(x\) as a set of steps, each consisting of three key components: 1) a title that identifies an aspect to be addressed in order to provide a complete and factual response,  2) a justification or reasoning that explains why this aspect is important and how it contributes to addressing the prompt, and 3) a query designed to gather information about the specified aspect from a corpus. To obtain a plan, we sample it from a planner model \( M_P \), which is an LLM guided by the plan generation prompt shown in Figure~\ref{fig:our-prompts} in Appendix~\ref{app:imp-details}. This prompt is designed to guide the LLM to analyze the input \( x \) and generate aspects that should be included in a complete and factual response. Next, using the queries specified in the generated plan \( p \), we employ the retrieval model \( R \) within a defined retrieval budget \( k \) to gather a supporting context. Specifically, for each component \( (a_i, q_i, r_i) \in p \), we retrieve \( \frac{k}{|p|} \) documents from the corpus \( C \). The resulting context for the plan \( p \) is denoted as \( I_p = \bigcup_{(a_i, q_i, r_i) \in p} R(q_i, \frac{k}{|p|}, C) \). To produce a response for the prompt \( x \), we leverage the generative model \( M_G \). This model, which is an LLM, takes the input \( x \), the generated plan \( p \), and the corresponding context \( I_p \) to generate a response. The process uses the "response generation with plan and context" prompt, as illustrated in Figure~\ref{fig:our-prompts} in Appendix~\ref{app:imp-details}. This prompt guides the model to incorporate the generated aspects and their associated reasoning from the plan, along with the provided context, to produce a comprehensive and factual response.

To sample a plan from \( M_p \), the most common strategy is greedy sampling, which returns the most probable plan for the given input \( x \). The most probable plan may not always yield the most complete and factual response. Alternatively, nucleus sampling \citep{nuc_sampling}, which introduces randomness, can generate diverse plans but risks reducing performance when only one plan is sampled. To balance these trade-offs, we propose sampling \( N \) plans using high-temperature, denoted as $P=\{p_i | p_i \sim M_P(x), \text{for}\,i=1, \dots, N\}$. This allows us to explore multiple strategies for answering \( x \), conducting a global search across the response space to identify diverse and potentially better plans for response generation. Finally, for each plan \( p \in P \), we use the generative model \( M_G \) to generate a response. This results in an initial set of responses, denoted as \( O_0 = \{M_G(x, p, I_p) \mid p \in P\} \), serving as the starting point for further refinement and selection in response to the input \( x \).

\paragraph{\textbf{Optimization.}}

We employ Self-Training \citep{rest, rest-em} as the optimization approach. Importantly, we only optimize the planner model \( M_P \), while keeping the generative model \( M_G \) frozen. For this purpose, for each input \( x \in D_{\text{train}} \), we sample \( B = 32 \) plans using a high temperature \( \tau = 0.7 \). We then generate a response for each plan and its corresponding context. To select high-quality plans that resulted in high-quality responses, we use the evaluation metric \( \mu \) and retain only plans that their corresponding responses achieved a score higher than the input-dependent threshold \( \alpha_x \), as follows:
\begin{equation*}
    D_{\text{plan}} = \{(x, p) \mid x \sim D_{\text{train}}; p \sim M_P(x); \mu(x, M_G(x, p, I_p)) \geq \alpha_x\}    
\end{equation*}
to form the training dataset $D_{\text{plan}}$ for training the the planner. We set the \( \alpha_x \) based on the score of the generated responses. Specifically, \( \alpha_x \) is chosen as the score corresponding to the top \( Z \)-percentile of the generated responses (we use $z=95$ by default, unless otherwise noted). This ensures that only the highest-scoring responses, as determined by the evaluation metric \( \mu \), are retained for training the planner model. Finally, we train the planner using a sequence-to-sequence loss function \citep{seq2seq} with $D_{\text{plan}}$, where for each $(x, p) \in D_{\text{plan}}$ the model to generates the plan $p$ as the output for the input \( x \).

\subsection{Local Exploitation through Refining}

While global exploration generates a diverse set of solutions, it often lacks the precision required to meet specific requirements. We observed that refining the response generated by a plan, using the same plan, leads to improved results. This suggests that focusing on enhancing existing solutions rather than exploring new ones can also yield to more accurate and complete responses. This iterative process of improving the response generated for a prompt \( x \) by a plan \( p \), using the same plan and refining the solution based on previous outputs, can be viewed as a local exploitation over the response space. Unlike global exploration, where both the plan and the output can vary, in local exploitation, the plan---the general instruction for the model in responding to the prompt---remains the same. It is the output that evolves through successive edits, refining the response according to the same guiding plan. To perform iterative refinement, we use the editing model \( M_E \), an LLM that uses the response editing prompt shown in Figure~\ref{fig:our-prompts} in Appendix~\ref{app:imp-details}. The model takes the input \( x \), the plan \( p \), and the previous output \( o_{t-1} \) generated using this plan as input, and produces the refined output \( o_t = M_E(x, p, o_{t-1}) \). This iterative process allows us to refine the initial set of responses \( O_0 \) from the global exploration phase. At each step, the updated responses are represented as \( O_t = \{M_E(x, p, o_{t-1}) \mid o_{t-1} \in O_{t-1}, p \in P\} \). By repeating this editing process \( T \) times, we obtain a final set of proposals, denoted as \( O_F = \bigcup_{t=0}^{T} O_t \), encompassing both the initial responses and the refined outputs after each editing step generated through the iterative steps. Therefore, the final response to the input \( x \) can be selected from the set of proposed responses \( O_F \), resulting from both global exploration using diverse plans and multiple rounds of local exploitation.

\paragraph{\textbf{Optimization.}}

To optimize the editing model \( M_E \), we sample a plan \( p \) from the optimized planner \( M_P \) for each input \( x \in D_{\text{train}} \). For each plan \( p \), we generate \( B = 8 \) pairs of outputs from the generative model \( M_G \) using a high sampling temperature \( \tau = 0.7 \). These pairs are selected such that the difference in their scores, as evaluated by the metric \( \mu \), is at least \( \beta \). This ensures that the training dataset for \( M_E \) has significant differences in responses, so that the model can learn how to improve the previous response. We form the training dataset $D_{\text{edit}}$ for training the editing model $M_E$ as:
\begin{align*}
    D_{\text{edit}} = \{(x, p, o_{0}, o_1) \mid x \sim D_{\text{train}}; p \sim M_P(x); o_0, o_1 \sim M_G(x, p, I_p); \mu(x, o_1) - \mu(x, o_0) \geq \beta\}
\end{align*}
where we set \( \beta = 0.1 \). To optimize \( M_E \), we use sequence-to-sequence loss \citep{seq2seq}. For each example \( (x, p, o_0, o_1) \in D_{\text{edit}} \), the model takes the input \( x \), the plan \( p \), and the lower-quality output \( o_0 \) as input and is trained to generate the higher-quality output \( o_1 \). This objective aligns the editing model's predictions with outputs that demonstrate improved quality, as defined by the evaluation metric \( \mu \).

\subsection{Response Selection through Ranking}

Previous steps produce a set of proposed responses \( O_F \), rather than a single response. To generate a final response \( o_f \), a selection mechanism is required to identify the most suitable response. For this, we use a reward model \( M_R \), which evaluates each candidate response based on the prompt \( x \) and assigns it a score between 0 and 1. To implement \( M_R \), we employ a text encoder model \( Enc \). The reward model computes the score as follows: $M_R(x, o) = \sigma(Enc([x.o]) \cdot W)$ where \( W \in \mathbb{R}^{d \times 1} \) is a trainable weight matrix, \( d \) represents the dimension of the encoder's output representations, \( \sigma \) is the sigmoid activation function, and $[.]$ is the concatenation with separate token function. This formulation allows \( M_R \) to evaluate the relevance and quality of a response \( o \) to the prompt \( x \), enabling the selection of the final response \( o_f \) as: $ o_f = {\mathrm{arg\,max}}_{o \in O_F} \, M_R(x, o)$.

\paragraph{\textbf{Optimization.}}

To optimize the reward model \( M_R \), we create a training dataset by sampling \( B = 8 \) pairs of plans \( p_0 \) and \( p_1 \) from the optimized planner \( M_P \) with a high temperature $\tau=0.7$ for each input \( x \in D_{\text{train}} \). The corresponding outputs \( o_0 = M_G(x, p_0, I_{p_0}) \) and \( o_1 = M_G(x, p_1, I_{p_1}) \), generated using the generative model \( M_G \), are included in the dataset if the difference in their scores \( \mu(x, o_1) - \mu(x, o_0) \) is at least \( \gamma \). Formally, the dataset for reward model is defined as:
\begin{align*}
    D_{\text{reward}} = \{(x, o_{0}, o_1) \mid x \sim D_{\text{train}}; p_0, p_1 \sim M_P(x);\\ 
    o_0\sim M_G(x, p_0, I_{p_0}); o_1\sim &M_G(x, p_1, I_{p_1});  \mu(x, o_1) - \mu(x, o_0) \geq \gamma\}
\end{align*}
where $\gamma=0.1$. To train $M_R$, following \citet{rlhf}, we minimize the following loss function:
\begin{equation*}
    L = \mathop{\mathbb{E}}_{(x, o_0, o_1) \sim D_{\text{reward}}} \left[ -\log(\sigma(M_R(x, o_1) - M_R(x, o_0))) \right]
\end{equation*}
where $\sigma$ is the sigmoid function.This pairwise loss function ensures that the reward model assigns higher scores to preferred outputs, as defined by the evaluation metric \( \mu \). This helps the model learn to distinguish response quality and align its predictions with the preferences encoded in \( \mu \).

\section{Experiments}

\subsection{Experimental Setup}

% \paragraph{\textbf{Dataset}}

% Our approach can be applied to any non-factoid question-answering dataset. For this study, we utilize the ANTIQUE \citep{antique} dataset, a retrieval-based dataset designed for non-factoid question answering. Unlike typical datasets, ANTIQUE does not include predefined gold responses to questions but provides a corpus containing the necessary information to answer them. The dataset consists of 2,426 training questions and 200 test questions. As a preprocessing step, we filter out documents with fewer than 50 words from the corpus to ensure the quality and richness of the documents used as the knowledge source. This document filtering process results in a corpus consisting of 97,327 documents.

\paragraph{\textbf{Datasets.}}

We use the ANTIQUE dataset \citep{antique}, a non-factoid QA benchmark with 2,426 training and 200 test questions. Since it lacks a validation set, we reserve 10\% of the training data for that purpose. The ANTIQUE's filtered corpus includes 97,327 documents (see Appendix~\ref{app:datasets} for details), which we use as the knowledge source for this dataset. We also utilize the TREC Web Track Diversity tasks from 2009 to 2012 \citep{trec09, trec10, trec11, trec12}. For the TREC Web Track Diversity tasks, no training set is available. The query set includes 200 queries, from which we remove navigational queries---those targeting specific webpages---resulting in 179 queries. We use the ClueWeb09 corpus \citep{clueweb09} as the document collection. Given the large size of ClueWeb09, we retrieve the top 1,000 documents per query using BM25 \citep{bm25}, following \citet{icat}, which results in a filtered corpus of 26,920 documents. The details about the datsets and filtering steps are provided in Appendix~\ref{app:datasets}.

% \paragraph{\textbf{Evaluation}}

% We evaluate the factuality and coverage of the generated responses using the ICAT \citep{icat} metric, which is specifically designed for this purpose.\footnote{ICAT has several variations, each requiring different levels of annotation. For our experiments, we use \( \text{ICAT}^{\text{A}} \), which automatically generates the topics that need to be covered for a query.} This metric leverages an LLM to generate subtopics related to the query, then decomposes the response into atomic claims to determine which subtopics are addressed. It also employs natural language inference (NLI) to fact-check the atomic claims. The final score is computed using the F1-measure, balancing the factuality of the response with its coverage of the subtopics. For more details, we refer the reader to \citet{icat}. As the LLM backbone, we follow \citet{icat} and use an instruction-tuned LLama 3.1 model with 8 billion parameters.\footnote{This model is available at: \url{https://hf.co/meta-llama/Llama-3.1-8B-Instruct}} For extracting atomic claims, we leverage the trained version of this model provided by ICAT.\footnote{Available at: \url{https://hf.co/algoprog/fact-generation-llama-3.1-8b-instruct-lora}} For NLI and fact verification, we employ a trained DeBERTa v3 \citep{he2023debertav3improvingdebertausing, Laurer_van_Atteveldt_Casas_Welbers_2024} model suggested by ICAT.\footnote{Available at: \url{https://hf.co/MoritzLaurer/DeBERTa-v3-base-mnli-fever-anli}} Lastly, for the knowledge source, we use the corpus of each dataset.

\paragraph{\textbf{Evaluation.}}

We evaluate the factuality and coverage using the ICAT metrics \citep{icat}, specifically designed for this purpose. ICAT offers three levels of annotation for evaluating comprehensiveness:  1) \( \text{ICAT}^{\text{M}} \): Requires a predefined set of subtopics for each query, along with annotations specifying which subtopics are addressed by each document in the corpus,  2) \( \text{ICAT}^{\text{S}} \): Similar to \( \text{ICAT}^{\text{M}} \), but leverages an LLM to determine which subtopics are covered by a document, eliminating the need for manual document-level annotations, and  3) \( \text{ICAT}^{\text{A}} \): Extends \( \text{ICAT}^{\text{S}} \) by using an LLM to generate the subtopics for a query, removing the dependency on predefined subtopic annotations.  ICAT also employs NLI to fact-check the claims in the generated response. The final score is calculated using the F-measure, balancing the factuality of the response with its coverage of the subtopics. For more details about ICAT, we refer the reader to \citet{icat}. Note that the \( \text{ICAT}^{\text{A}} \), which shows the highest correlation with human judgment, relies on an LLM to generate subtopics that responses are expected to cover. To ensure a fair evaluation, we use a different LLM within \ourmethod than the one employed for ICAT. The configuration of ICAT used in this paper is detailed in Appendix~\ref{app:eval}.

\paragraph{\textbf{\ourmethod Configurations.}}

We use the instruction-tuned Gemma 2 \citep{gemmav2} with 2.6 billion parameters as the LLM and ModernBERT-base \citep{modern-bert} with 150 million parameters as the reward model. We set the maximum input and output length to 4096 tokens. For sampling from the generative model, we use nucleus sampling \citep{nuc_sampling} with a temperature of \( \tau = 0.1 \). For the editing model, nucleus sampling is applied with \( \tau = 0 \). When sampling plans with the planner, we use a temperature of \( \tau = 0.7 \) for global exploration and \( \tau = 0 \) otherwise. We define the exploration budget as the total number of responses generated and edited during the process of responding to an input.\footnote{\ourmethod's average generated output length in our experiments is \(316.4 \pm 144.7\) words.} We perform \( N = 4 \) global and \( T = 4 \) local exploitation steps to achieve a generation budget of 16, unless stated otherwise. As a retriever, we use a BERT model\footnote{Available at: \url{https://hf.co/Snowflake/snowflake-arctic-embed-l}} \citep{bert} pre-trained on retrieval tasks \citep{retriever-snowflake} to retrieve \( k=40 \) for ANTIQUE and \( k=5 \) for TREC datasets. The details for training are provided in Appendix~\ref{app:experiment-setup}.

\paragraph{\textbf{Baselines.}}

We use a range of baselines, including open-source and proprietary. As open-source, we utilize LLama 3.2 \citep{llama3} with 1.2B, Gemma 2 \citep{gemmav2} with 2.6B, and Phi 3 \citep{phi3} with 3.8B parameters. They are used with and without RAG and CoT. Additionally, we introduce baselines using best-of-N for each backbone, maintaining the same computational budget as \ourmethod. We also train Gemma 2 with RAG, the same backbone used in \ourmethod, via self-training with ICAT as the reward model, providing a trained baseline for comparison with \ourmethod. Finally, we apply Maximal Marginal Relevance (MMR) \citep{mmr} to re-rank the top 1,000 documents retrieved by the retriever, investigating whether diverse retrieval results can enhance the coverage of generated responses. As proprietary, we use two capable models with strong reasoning: GPT-4o-mini \citep{gpt4o} and Gemini 2 Flash \citep{gemini2}. These models naturally perform CoT; we do not explicitly prompt them for it. Additionally, due to the high cost of the Best-of-N, we do not apply this to them. The details about the baselines are provided in Appendix~\ref{app:baseline}.

\subsection{Main Findings}
\label{sec:main-findings}

\begin{wraptable}{!tr}{0.5\textwidth}
    \centering
    \vspace{-0.3cm}
    \caption{Performance on ANTIQUE using ICAT-A. The $\dagger$ and $\ddagger$ show statistically significant improvements over the best open-source and proprietary baselines, respectively, using t-test ($p < 0.05$).}
    % \vspace{-0.4cm}
    \label{tab:main-results}
    \adjustbox{max width=0.5\textwidth}{
    \begin{tabular}{cp{3.4cm}|c|c||c}
        \toprule
        & \multirow{2}{*}{\textbf{Method}} & \multirow{2}{*}{\textbf{$\text{ICAT}_{\text{Coverage}}$}} & \multirow{2}{*}{\textbf{$\text{ICAT}_{\text{Factuality}}$}} & \multirow{2}{*}{\textbf{$\text{ICAT-A}_{\text{1}}$}} \\
        & & & &  \\
        \midrule
        \multicolumn{5}{c}{\textbf{Proprietary LLMs}} \\
        \midrule
        1 & Gemini 2.0 Flash & \textbf{0.7057} & 0.4488 & 0.5214  \\
        2 & GPT-4o mini & 0.6551 & 0.4934 & 0.5376  \\
        \midrule
        \multicolumn{5}{c}{\textbf{Retrieval-Augmented Proprietary LLMs}} \\
        \midrule
        3 & RAG Gemini 2.0 Flash & 0.6499 & 0.5474 & 0.5640 \\
        4 & RAG GPT-4o mini &  0.6439 & 0.5354 & 0.5576 \\
        \midrule
        \multicolumn{5}{c}{\textbf{Open-Source LLMs}} \\
        \midrule
        5 & Llama 3.2 & 0.3959 & 0.3201 & 0.3251 \\
        6 & \quad - w/ CoT & 0.3523 & 0.3444 & 0.3207 \\
        % \quad - w/ Best-of-N (using our $M_R$) & 0.5210 & 0.4288 & 0.4311 & - & - \\
        7 & \quad - w/ Best-of-N & 0.4521 & 0.3995 & 0.3924 \\
        \midrule
        8 & Phi 3 mini & 0.5483 & 0.4433 & 0.4511 \\
        9 & \quad - w/ CoT & 0.4973 & 0.4219 & 0.4116 \\
        % \quad - w/ Best-of-N (using our $M_R$) & 0.6065 & 0.4924 & 0.5020 & - & - \\
        10 & \quad - w/ Best-of-N & 0.5489 & 0.4754 & 0.4741 \\
        \midrule
        11 & Gemma 2 & 0.6064 & 0.4936 & 0.5143 \\
        12 & \quad - w/ CoT & 0.5257 & 0.4890 & 0.4659 \\
        % \quad - w/ Best-of-N (using our $M_R$) & 0.6230 & 0.5055 & 0.5268 & - & - \\
        13 & \quad - w/ Best-of-N & 0.5789 & 0.4787 & 0.4952 \\
        14 & \quad - w/ Self-Training & 0.5839 & 0.5268 & 0.5243 \\
        
        \midrule
        \multicolumn{5}{c}{\textbf{Retrieval-Augmented Open-Source LLMs}} \\
        \midrule
        15 & RAG Llama 3.2 & 0.3162 & 0.3295 & 0.2872 \\
        16 & \quad - w/ CoT & 0.3112 & 0.3243 & 0.2878 \\
        % \quad - w/ Best-of-N (using our $M_R$) & 0.5093 & 0.5366 & 0.4830 & - & - \\
        17 & \quad - w/ Best-of-N & 0.3564 & 0.3712 & 0.3363 \\
        18 & \quad - MMR Reranking & 0.3005 & 0.2830 & 0.2751 \\
        \midrule
        19 & RAG Phi 3 mini & 0.5369 & 0.5557 & 0.5022 \\
        20 & \quad - w/ CoT & 0.5173 & 0.5635 & 0.5071 \\
        % \quad - w/ Best-of-N (using our $M_R$) & 0.6143 & 0.5792 & 0.5630 & - & - \\
        21 & \quad - w/ Best-of-N & 0.5493 & 0.5386 & 0.5021 \\
        22 & \quad - MMR Reranking &  0.5541 & 0.5656 & 0.4758 \\
        \midrule
        23 & RAG Gemma 2 & 0.5457 & 0.5904 & 0.5256 \\
        24 & \quad - w/ CoT & 0.5028 & 0.5655 & 0.4880 \\
        25 & \quad - w/ Best-of-N & 0.4873 & 0.5809 & 0.4901 \\
        26 & \quad - w/ Self-Training & 0.5382 & 0.6054 & 0.5310 \\
        27 & \quad - MMR Reranking & 0.5162 & 0.5977 & 0.5006 \\
        \midrule
        \midrule
        28 & \textbf{\ourmethod} & 0.6318$^{\dagger}$ & \textbf{0.6237$^{\ddagger}$} & \textbf{0.6010$^{\dagger \ddagger}$} \\
        \midrule
        29 & \quad - w/o Global & 0.6423$^{\dagger}$ & 0.6073$^{\ddagger}$ & 0.5961$^{\dagger \ddagger}$ \\

        30 & \quad - w/o Local & {0.6554$^{\dagger}$} & 0.6017$^{\ddagger}$ & 0.5960$^{\dagger \ddagger}$ \\

        31 & \quad - w/o Local \& Global & 0.6543$^{\dagger}$ & 0.5808$^{\ddagger}$ & 0.5832$^{\dagger}$ \\

        32 & \quad - w/o Local \& Global & \multirow{2}{*}{0.6318$^{\dagger}$} & \multirow{2}{*}{0.5512} & \multirow{2}{*}{0.5556} \\
        & \quad \& Self-Training & & & \\

        % \quad - w/o Local \& Global Exploration \& Planning & & &  & \color{PineGreen}{+0\% $\uparrow$} & \color{Maroon}{-0\% $\downarrow$} \\
        \bottomrule
    \end{tabular}}
    \vspace{-0.6cm}
\end{wraptable}

\paragraph{\textbf{How does \ourmethod perform compared to baselines?}}

We compare \ourmethod against different baselines. The results of these experiments on the ANTIQUE dataset are presented in Table~\ref{tab:main-results}. These results demonstrate that \ourmethod statistically significantly outperforms both open-source and proprietary LLMs on the ICAT-A metric, emphasizing its superior performance in generating complete and factual responses. Specifically, \ourmethod achieves a 13.1\% relative improvement over the best open-source baseline (row 26 in Table~\ref{tab:main-results}) and a 6.5\% improvement over the best proprietary baseline (row 3). This highlights the effectiveness of \ourmethod in improving factuality, coverage, and their aggregation (ICAT-A).

The results in Table~\ref{tab:main-results} suggest that RAG enhances the performance of LLMs with generating more factual responses by incorporating relevant retrieved documents. However, it may lead to a reduction in coverage, as the retrieved documents tend to be similar to one another, limiting the coverage. Another observation is that the CoT tends to negatively affect the performance of LLMs in most cases. This occurs because LLMs are typically trained to apply CoT for reasoning and mathematical tasks, inherently different generating factual and complete responses. Thus, CoT may not be as effective for this task. The Best-of-N approach generally enhances the performance of LLMs, remaining less effective for the Gemma 2. Moreover, self-training proves to be the most effective strategy for training baselines, though it still significantly lags behind \ourmethod in overall performance (row 26 vs 28). Finally, we find that using MMR to diversify the retrieval results does not yield improvement in coverage and factuality in most cases; instead, it leads to a drop in performance (rows 18, 22, and 27). 
% Given this, \ourmethod shows the best and most promising results for this task.

\paragraph{\textbf{How do global and local exploration affect performance?}}

We evaluate global and local exploitation separately, each using the same budget as \ourmethod with both combined (i.e., 16 generations). On ANTIQUE, we conduct experiments using only local exploitation, where a single plan is sampled greedily (with a temperature of \(\tau = 0.0\)) and refined through 16 editing steps, and only global exploration, where 16 plans are sampled using a higher temperature of \(\tau = 0.7\) from the planner. The results are reported in Table~\ref{tab:main-results} (row 29 for local exploitation only and row 30 for global exploration only). The findings indicate that while using either local or global exploration achieves nearly identical ICAT-A scores, both are suboptimal compared to combining both. However, both methods outperform the planning-only with (row 31) and without (row 32) self-training, achieving statistically significant improvements over all baselines. These results highlight the effectiveness of global and local exploitation and demonstrate their complementary strengths when combined in \ourmethod.

% \begin{table}
%     \centering
%     \caption{Performance of \ourmethod without self-training, local and global exploration compared to RAG baselines on TREC. The $\dagger$ shows statistically significant improvements over the best baseline, as determined by a t-test ($p < 0.05$).}
%     \vspace{-0.4cm}
%     \label{tab:results-trec}
%     \adjustbox{max width=\linewidth}{
%     \begin{tabular}{cp{3cm}|c|c||c}
%         & \multirow{2}{*}{\textbf{Method}} & \multirow{2}{*}{\textbf{$\text{ICAT}_{\text{Coverage}}$}} & \multirow{2}{*}{\textbf{$\text{ICAT}_{\text{Factuality}}$}} & \multirow{2}{*}{\textbf{$\text{ICAT-A}_{\text{1}}$}} \\
%         & & & &  \\
%         \hline
%         1 & RAG Llama 3.2 & 0.1568 & 0.1699 & 0.1476 \\
%         2 & RAG Phi mini & 0.3344 & 0.6597 & 0.3380 \\
%         3 & RAG Gemma 2 & 0.5079 & \textbf{0.6720} & 0.5148 \\
%         \hline
%         \hline
%         4 & \ourmethod & \textbf{0.6665$^\dagger$} & 0.6325 & \textbf{0.5943$^\dagger$} \\
%     \hline
%     \end{tabular}}
% \end{table}

\paragraph{\textbf{How does planning alone with and without self-training affect performance?}}

We focus on evaluating the planner without any exploration. We sample a single plan greedily (\(\tau = 0.0\)) to generate responses. We test both the untrained and self-trained planner. The results in Table~\ref{tab:main-results} on ANTIQUE show that the self-trained planner (row 31) alone is suboptimal compared to \ourmethod, but it achieves a $4.9\%$ relative improvement in ICAT-A compared to the untrained planner (row 32). This shows the effectiveness of self-training in improving the planner’s ability to generate better plans. The untrained planner (row 32) still outperforms the best-performing open-source baseline (row 26) with a $4.6\%$ relative improvement, showing the value of planning even without training or exploration. To explore it further, we compare \ourmethod without self-training, local, and global exploration to the best RAG baseline on the TREC dataset, which does not include a training set. Since the TREC dataset includes human annotations for subtopics that need to be covered for each query, we report all variations of the ICAT on TREC. As reported in Table~\ref{tab:results-trec}, \ourmethod with a untrained planner and no exploration achieves a statistically significant improvement over the baseline, with a $9.4\%$, $36.3\%$, and $15.4\%$ relative gain on ICAT-M, ICAT-S, and ICAT-A, respectively. This shows that \ourmethod significantly improves performance across different levels of annotated data availability.

\paragraph{\textbf{How does planner's self-training threshold affect performance?}}

\begin{figure}
    \centering
    \includegraphics[width=\linewidth]{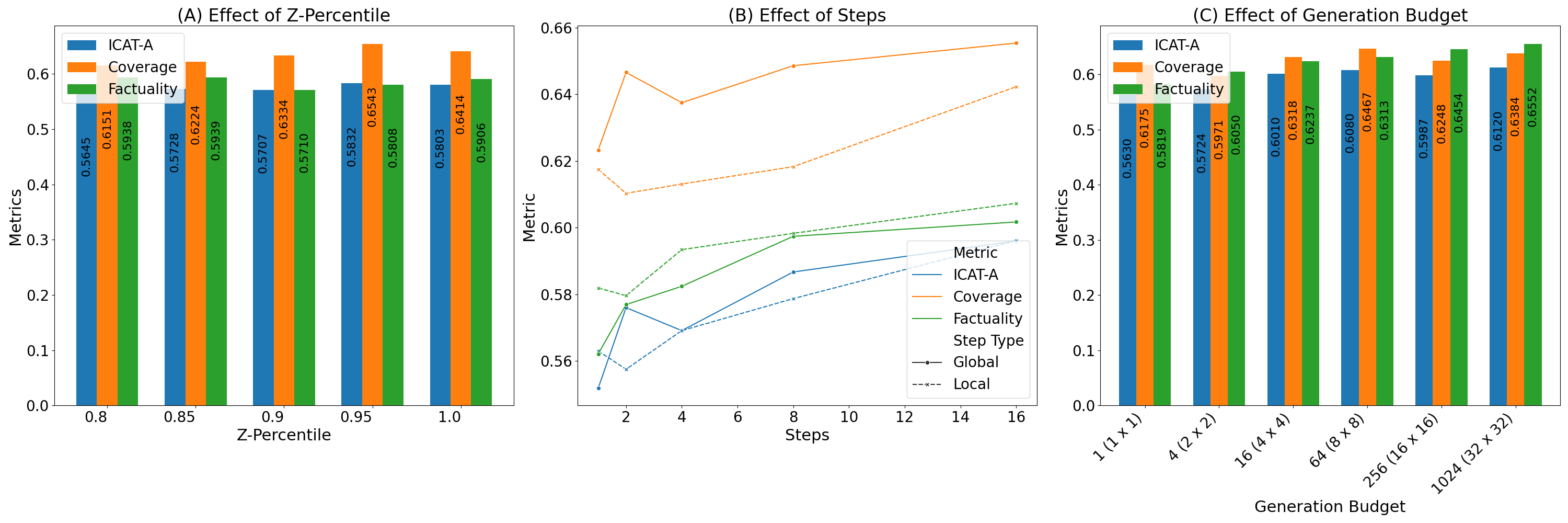}
    \vspace{-0.8cm}
    \caption{Effect of (A) threshold ($Z$) for planner training, (B) local/global steps, and (C) total budget on \ourmethod's performance on ANTIQUE. Larger versions appear in Figures~\ref{fig:self-training-percentile}, \ref{fig:local-global-steps}, and \ref{fig:gen-budget} in Appendix~\ref{app:figures}.}
    \label{fig:ablation}
    \vspace{-0.4cm}
\end{figure}

An important hyperparameter in \ourmethod for training the planner is the top Z-Percentile of the generated plans to be used for training. We train the planner using different values for \( Z \) and evaluate the planner with them on the ANTIQUE dataset. We generate a single plan greedily ($\tau = 0.0$) and produce a response. The results in Figure~\ref{fig:ablation} (A) indicate that as \( Z \) increases, the results improve, as the model is trained on higher-quality plans. The best performance occurs at \( Z = 0.95 \). However, setting \( Z = 1 \), i.e, only the output with highest score being selected for training, leads to missing high-quality outputs that could aid training.

\paragraph{\textbf{How does global and local exploitation budget affect performance?}}

Here, we conduct experiments on ANTIQUE using only local exploitation with a single plan sampled greedily ($\tau = 0.0$) and edited $N$ times, and only global exploration with $N$ plans sampled using a temperature of $\tau = 0.7$ from the planner. The results in Figure~\ref{fig:ablation} (B) show that increasing the number of steps leads to improvements across all aspects, with the ICAT-A metric being nearly identical for both with 16 generated outputs. However, it can be observed that increasing the number of global plans results in higher coverage, while increasing local exploitation steps leads to higher factuality. This indicates that sampling multiple plans produces outputs that cover more topics, but may lack factual accuracy. In contrast, sampling a single plan and applying multiple local editing steps results in lower coverage but higher factual accuracy. Given this, we show that the primary contribution of global exploration is to enhance coverage, while the main contribution of local exploitation is to improve factuality.

\paragraph{\textbf{How does exploration budget affect performance?}}

We evaluate \ourmethod on the ANTIQUE dataset under different budgets: 1, 4, 16, 64, 256, and 1024 responses per input, allocated equally to global and local exploitation. The results in Figure~\ref{fig:ablation} (C) show that increasing the budget leads to improved performance on the ICAT-A. A general trend of improvement in coverage is also observed, though with some fluctuations. Factuality shows a consistent increase as the budget grows. This indicate that larger exploration budgets improve factuality and topic coverage, with a stronger impact on factuality.

\begin{table*}
    \centering
    \caption{Performance of \ourmethod w/o self-training and exploration compared to baseline on TREC using variations of ICAT. The $\dagger$ shows statistically significant improvements using t-test ($p < 0.05$).}
    % \vspace{-0.4cm}
    \label{tab:results-trec}
    \adjustbox{max width=\textwidth}{
    \begin{tabular}{l|c|cc|cc|cc}
        \toprule
        \multirow{2}{*}{\textbf{Metric}} & \multirow{2}{*}{\textbf{Factuality}} & \multicolumn{2}{c}{\textbf{Manual}} & \multicolumn{2}{|c}{\textbf{Semi-Automatic}} & \multicolumn{2}{|c}{\textbf{Automatic}} \\
        \cmidrule{3-8}
        & & \textbf{Coverage} & \textbf{$\text{ICAT-M}_{\text{1}}$} & \textbf{Coverage} & \textbf{$\text{ICAT-S}_{\text{1}}$} & \textbf{Coverage} & \textbf{$\text{ICAT-A}_{\text{1}}$} \\
        \midrule
        RAG Gemma 2 & \textbf{0.6720} & 0.2980 & 0.3203 & 0.1970 & 0.2294 & 0.5079 & 0.5148 \\
        \ourmethod (w/o self-training \& exploration) & 0.6325 & \textbf{0.3523$^\dagger$} & \textbf{0.3507$^\dagger$} & \textbf{0.2819$^\dagger$} & \textbf{0.3129$^\dagger$} & \textbf{0.6665$^\dagger$} & \textbf{0.5943$^\dagger$} \\
        \bottomrule
    \end{tabular}}
    \vspace{-0.4cm}
\end{table*}

\begin{wraptable}{!r}{0.4\textwidth}
    \centering
    \vspace{-0.7cm}
    \caption{Human alignment (\%) of \ourmethod and \textit{RAG Gemma 2 w/ Self-Training}.}
    \label{tab:human-anot}
    \adjustbox{max width = 0.4\textwidth}{
    \begin{tabular}{l|c|c|c}
        \toprule
        \textbf{Winner} & \textbf{Coverage} & \textbf{Factuality} & \textbf{Overall}  \\
        \midrule
         \ourmethod & \textbf{64} & 35 & \textbf{63} \\
         Baseline & 26 & 9 & 29 \\
         Tie & 10 & \textbf{56} & 8 \\
         \bottomrule
    \end{tabular}}
    \vspace{-0.2cm}
\end{wraptable}

\paragraph{\textbf{How does \ourmethod align with human preferences?}}

We randomly selected 50 queries from the ANTIQUE dataset and generated outputs using \ourmethod and \textit{RAG Gemma 2 w/ Self-Training} due to its strong performance and the fact that it uses the same LLM as \ourmethod for a fair comparison. Two annotators evaluated the outputs based on three criteria: coverage of topics, factual accuracy, and overall quality of responses. The inter-annotator agreement with Cohen's $\kappa$ is 0.6189. The results are presented in Table~\ref{tab:human-anot}. In coverage, annotators preferred \ourmethod 64\% of cases, compared to 26\% for the baseline. In factuality, the outputs of both models were rated equally in 56\% of cases, but in the remaining, \ourmethod was preferred 35\%, while the baseline was chosen in only 9\% of cases. Overall, \ourmethod was selected as the preferred output in 63\% of cases, compared to 29\% for the baseline. This show that \ourmethod aligns more with human preferences. A case study of responses generated by \ourmethod is shown in Appendix~\ref{app:case-study}.

\section{Conclusion}

We introduce \ourmethod, an approach for improving factuality and coverage of  LLM's generated responses. \ourmethod begins by generating a diverse set of plans for responding to a prompt and retrieves information from a knowledge source to gather the necessary information for executing each plan. It then generates a response for each plan and iteratively refines them to enhance their factual coverage. Finally, a reward model selects the most factual and complete response from the set of generated proposals. Experiments on the ANTIQUE and TREC datasets show that \ourmethod outperforms both open and proprietary baselines by up to a 13.1\% and 6.5\% improvement, respectively. Human evaluation reveals that \ourmethod has considerably higher agreement with human preferences compared to baselines.

\bibliographystyle{plainnat}
\bibliography{custom}

\appendix

\section{The \ourmethod Framework Implementation Details}
\label{app:imp-details}

\begin{figure}[!ht]
    \centering
    \includegraphics[width=\linewidth]{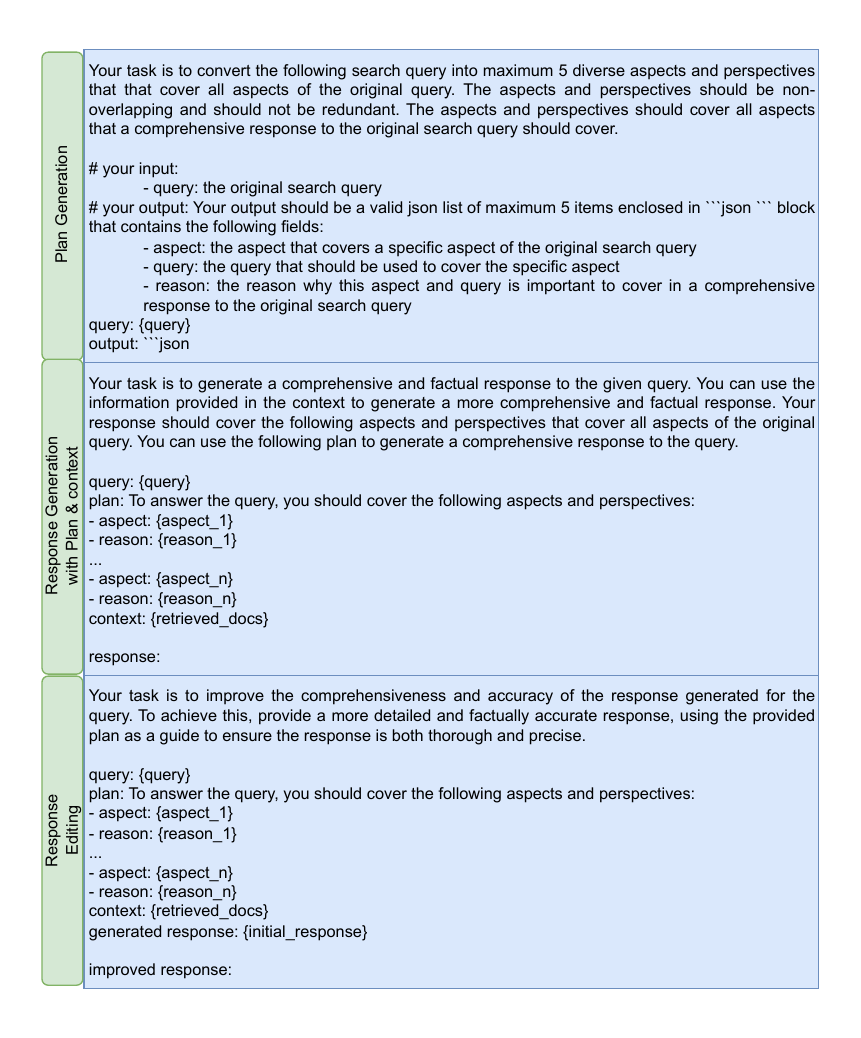}
    \vspace{-0.6cm}
    \caption{The prompt templates used with different components in the \ourmethod framework.}
    % \vspace{-0.6cm}
    \label{fig:our-prompts}
\end{figure}

\ourmethod comprises three main components: (1) a planner, which generates a structured plan consisting of the key aspects to be addressed in the response, the rationale for including each aspect, and corresponding retrieval queries to gather relevant information; (2) a generative model, which produces an initial response conditioned on the input question, the generated plan, and the retrieved evidence; and (3) an editing model, which refines the initial response to improve its overall quality. The prompt templates used for each component are illustrated in Figure~\ref{fig:our-prompts}.

To generate the plan, we prompt the LLM to produce output in a structured JSON format. In rare instances---fewer than 0.1\% of cases in all of our experiments---the model may fail to produce a valid JSON output. When this occurs, we re-prompt the model, incrementally increasing the decoding temperature with each attempt until it reaches a maximum of 1. In practice, this issue is infrequent and does not pose a significant challenge to the overall workflow.

\section{Datasets \& Corpus}
\label{app:datasets}

We use ANTIQUE \citep{antique}, a retrieval dataset designed for non-factoid question answering, and TREC Web Track Diversity tasks from 2009 to 2012 \citep{trec09, trec10, trec11, trec12}. These datasets do not include predefined gold responses to questions, but provide a corpus containing the necessary information to answer them. It is important to note that the recently introduced TREC RAG track \citep{pradeep2024initialnuggetevaluationresults} has proposed the concept of Nugget evaluation for assessing coverage in responses. However, since the judgments are not publicly accessible yet, we do not use them.

The ANTIQUE dataset consists of 2,426 training questions and 200 test questions. As a pre-processing step, we filter out documents with fewer than 50 words from the corpus to ensure the quality and richness of the documents used as the knowledge source. This document filtering process results in a corpus consisting of 97,327 documents. 

For the TREC Web Track Diversity tasks, there is no training dataset available, but the query set consists of 200 queries. We exclude queries that seek information about a specific webpage (navigational), reducing the set to 179 queries. For the corpus, we use the ClueWeb09 corpus \citep{clueweb09}. Note that we only use this dataset to evaluate the \ourmethod framework under the zero-shot setting, as it does not include any training query set. 

\section{Evaluation Metric Details}
\label{app:eval}

We evaluate the factuality and coverage of the generated responses using the ICAT metric \citep{icat}, which is specifically designed for this purpose. ICAT offers three levels of annotation for evaluating responses:  1) \( \text{ICAT}^{\text{M}} \): Requires a predefined set of subtopics for each query, along with annotations specifying which subtopics are addressed by each document in the corpus,  2) \( \text{ICAT}^{\text{S}} \): Similar to \( \text{ICAT}^{\text{M}} \), but leverages an LLM to determine which subtopics are covered by a document, eliminating the need for manual document-level annotations, and  3) \( \text{ICAT}^{\text{A}} \): Extends \( \text{ICAT}^{\text{S}} \) by using an LLM to generate the subtopics for a query, removing the dependency on predefined subtopic annotations.  ICAT also employs natural language inference (NLI) to fact-check the claims in the generated response. The final score is calculated using the F-measure, balancing the factuality of the response with its coverage of the subtopics. For more details, we refer the reader to \citet{icat}. 

For the LM backbone, we follow \citet{icat} and use an instruction-tuned LLama 3.1 model with 8 billion parameters.\footnote{Available at: \url{https://hf.co/meta-llama/Llama-3.1-8B-Instruct}} For extracting atomic claims, we leverage the trained version of this model provided by ICAT.\footnote{Available at: \url{https://hf.co/algoprog/fact-generation-llama-3.1-8b-instruct-lora}} For NLI and fact verification, we employ a trained DeBERTa v3 \citep{he2023debertav3improvingdebertausing, Laurer_van_Atteveldt_Casas_Welbers_2024} model suggested by ICAT.\footnote{Available at: \url{https://hf.co/MoritzLaurer/DeBERTa-v3-base-mnli-fever-anli}} As the knowledge source, we use the corresponding corpus in each of the evaluation datasets, i.e., the ANTIQUE corpus and the ClueWeb09-Category B English corpus for the TREC Web Track queries. Spam documents were removed from the ClueWeb corpus using the Waterloo Spam Scorer with the 70\% threshold.\footnote{Available at \url{https://plg.uwaterloo.ca/~gvcormac/clueweb09spam/}.}

\section{Experimental Setup}
\label{app:experiment-setup}

We use the Adam optimizer \citep{adam} with a learning rate of \( 5 \times 10^{-5} \) for training the LLMs and \( 1 \times 10^{-5} \) for training the reward model. Gradient clipping is applied with a value of 1, and the training is conducted for a maximum of 2000 steps. A warmup phase is set for 2.5\% of the training steps, following a linear learning rate scheduler. Models are evaluated every 100 steps using 10\% of the training set as a randomly sampled validation subset, and the checkpoint with the best performance is selected. We set the combined maximum input and output length to 4096 tokens. We use the instruction-tuned Gemma 2 \citep{gemmav2} with 2.6 billion parameters as the LLM and ModernBERT-base \citep{modern-bert} with 150 million parameters as the reward model. The batch size for all experiments is set to 64. 

Experiments use 4 NVIDIA A100 GPUs (80GB VRAM) and 128GB of RAM. For sampling from the generative model \( M_G \), we use nucleus sampling \citep{nuc_sampling} with a temperature of \( \tau = 0.1 \). For the editing model \( M_E \), nucleus sampling is applied with \( \tau = 0 \). When sampling plans with the planner $M_P$, we use a nucleus sampling temperature of \( \tau = 0.7 \) for global exploration and \( \tau = 0 \) otherwise. We define the exploration budget as the total number of responses generated and edited during the process of responding to an input.\footnote{\ourmethod's average generated output length in our experiments is \(316.4 \pm 144.7\) words.} We perform \( N = 4 \) global and \( T = 4 \) local exploitation steps to achieve a total generation budget of 16, unless stated otherwise. As a retriever, we use a BERT model\footnote{Available at: \url{https://hf.co/Snowflake/snowflake-arctic-embed-l}} \citep{bert} pre-trained on retrieval tasks \citep{retriever-snowflake}. For indexing, we employ the Faiss library \citep{faiss} to construct a hybrid IVF-HNSW index \citep{hnsw} for ANTIQUE and a flat index for TREC, chosen based on the corpus size. The total retrieval budget for \ourmethod is set to \( k=40 \) for the ANTIQUE dataset and \( k=5 \) for the TREC dataset. These are chosen based on the document length in each corpus and the context size of the LLMs.

\section{Baselines}
\label{app:baseline}

\begin{figure}
    \centering
    \includegraphics[width=0.9\linewidth]{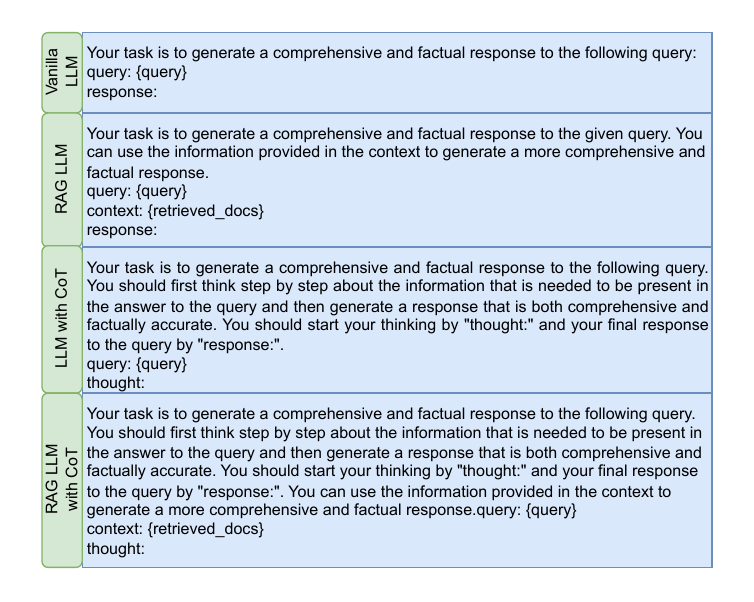}
    \vspace{-0.6cm}
    \caption{The prompts used by the baselines.}
    \vspace{-0.4cm}
    \label{fig:baseline-prompts}
\end{figure}

We leverage a variety of baseline LLMs of different sizes, both open-source and proprietary, with and without retrieval augmentation. The prompts used for the baselines are provided in Figure~\ref{fig:baseline-prompts}. For retrieval augmentation, we use the same retriever \ourmethod. For each baseline, we set the retrieval budget based on the performance on the validation set, ranging between 10 and 40, similar to the configuration used for \ourmethod. These baselines include:
\begin{itemize}[leftmargin=1em]
    \item \textbf{Open-Source:} We utilize three open-source instruction-tuned LLMs as the backbone for baselines: LLama 3.2 \citep{llama3}, with 1.2 billion parameters,\footnote{Available at: \url{https://hf.co/meta-llama/Llama-3.2-1B-Instruct}} Gemma 2 \citep{gemmav2}, with 2.6 billion parameters,\footnote{Available at: \url{https://hf.co/google/gemma-2-2b-it}} and Phi 3 \citep{phi3}, with 3.8 billion parameters.\footnote{Available at: \url{https://hf.co/microsoft/Phi-3-mini-4k-instruct}} For CoT models, we evaluate only the final response and do not assess the intermediate reasoning steps. For Best-of-N, we generate \( N = 16 \) outputs for each LLM with a temperature of $0.7$ using nucleus sampling \citep{nuc_sampling}, rerank them using an off-the-shelf reranking model,\footnote{Available at: \url{https://hf.co/cross-encoder/ms-marco-MiniLM-L-12-v2}} and select the top-ranked output as the final response. We also train Gemma 2 using self-training with ICAT as the reward model, in the same setting as \ourmethod. We leverage the high-scoring outputs of the model to train the model, enabling it to learn how to generate similar high-quality responses. This allows us to assess the potential improvements self-training can contribute to baseline models. Finally, we employ Maximal Marginal Relevance (MMR) \citep{mmr} with $\lambda = 0.1$ to rerank the top 1,000 documents retrieved by the retriever, investigating whether diverse retrieval results can enhance coverage of the generated responses.

    \item \textbf{Proprietary:} For proprietary LLMs, we use two highly capable models with strong reasoning abilities: GPT-4o-mini\footnote{Available at: \url{https://platform.openai.com/docs/models/gpt-4o-mini}} from OpenAI and Gemini 2 Flash\footnote{Available at: \url{https://ai.google.dev/gemini-api/docs/models/gemini-v2}} from Google. These models inherently perform CoT, so we do not explicitly prompt them for this. Additionally, due to the high cost associated with the Best-of-N approach, we do not apply this method to the proprietary LLMs.
\end{itemize}

\section{Figures with More Details}
\label{app:figures}

\begin{figure}
    \centering
    \includegraphics[width=0.9\linewidth]{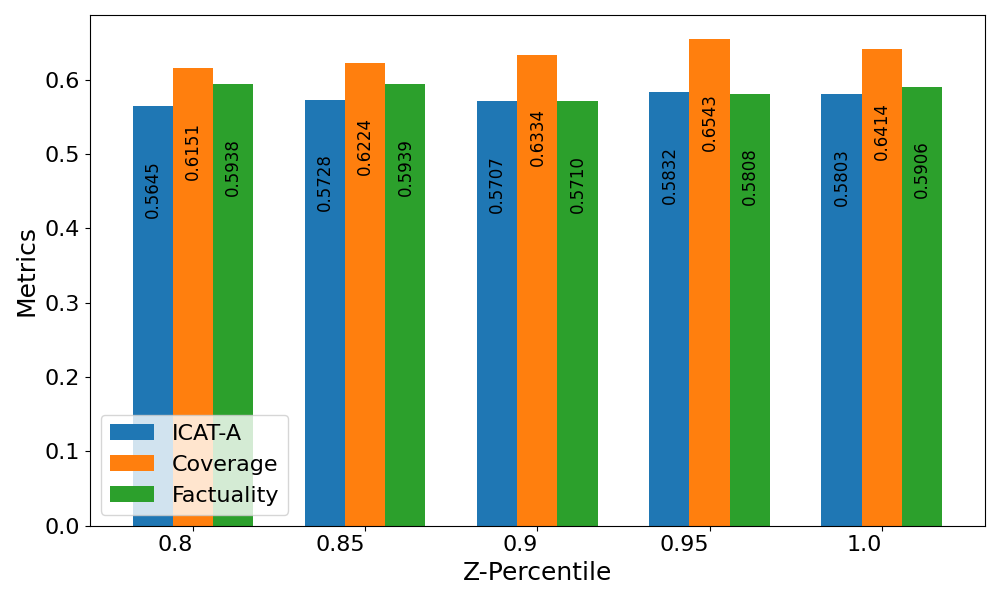}
    \vspace{-0.4cm}
    \caption{Effect of generated plan selection threshold for self-training planner on performance for the ANTIQUE dataset.}
    \label{fig:self-training-percentile}
    \vspace{-0.4cm}
\end{figure}

\begin{figure}
    \centering
    \includegraphics[width=0.9\linewidth]{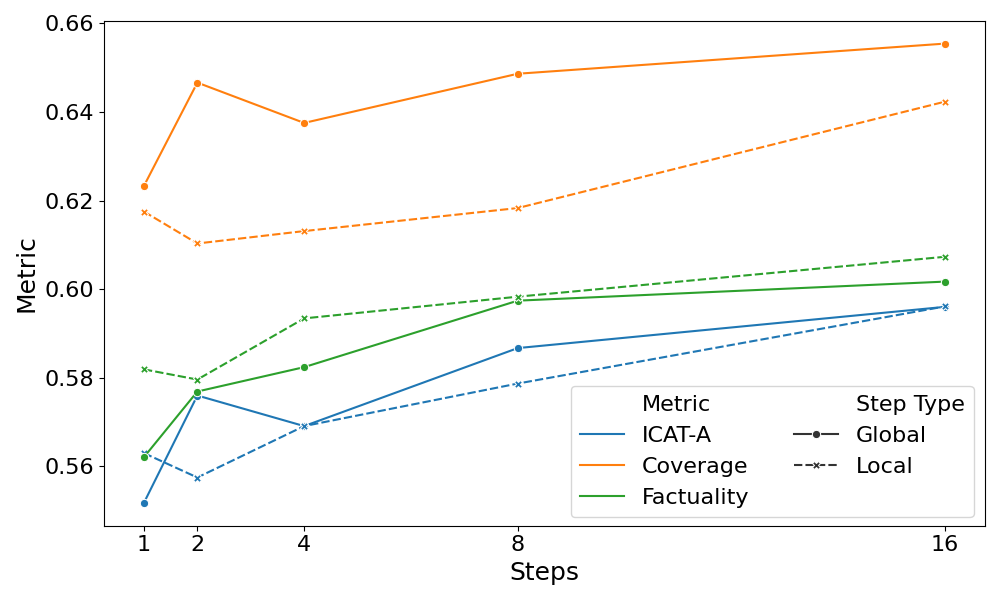}
    \vspace{-0.4cm}
    \caption{Effect of local and global exploration steps on \ourmethod's performance for the ANTIQUE dataset.}
    \vspace{-0.4cm}
    \label{fig:local-global-steps}
\end{figure}

\begin{figure}
    \centering
    \includegraphics[width=0.9\linewidth]{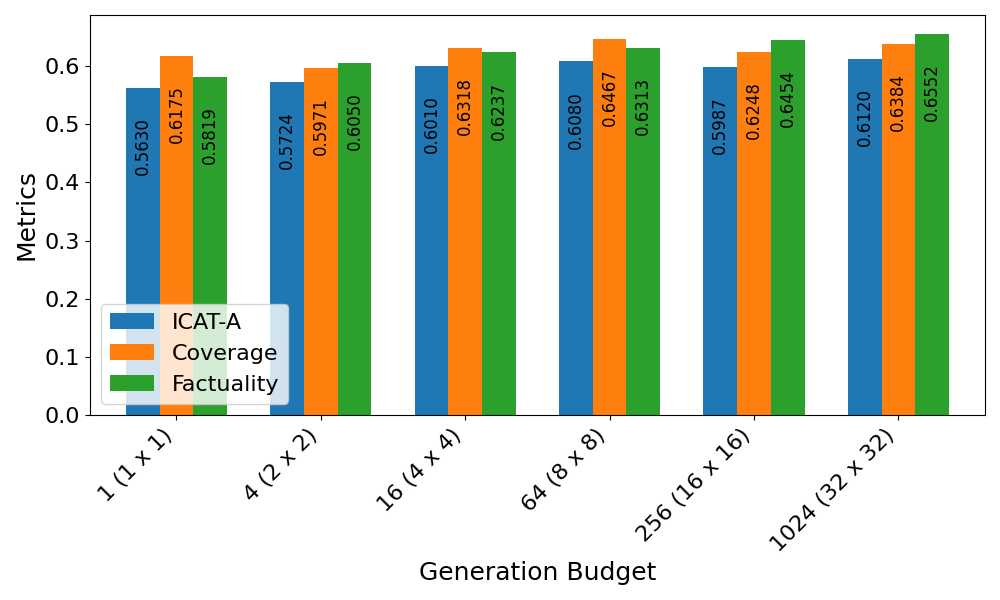}
    \vspace{-0.4cm}
    \caption{Effect of generation budget on the performance of \ourmethod on the ANTIQUE dataset.}
    \label{fig:gen-budget}
    \vspace{-0.4cm}
\end{figure}

Figure~\ref{fig:self-training-percentile} addresses the question: How does the planner's self-training threshold affect performance? A detailed explanation of the results can be found in Section~\ref{sec:main-findings}. Figure~\ref{fig:local-global-steps} answers the question: How does the global and local exploitation budget affect performance? A detailed explanation of the results is provided in Section~\ref{sec:main-findings}. Figure~\ref{fig:gen-budget} answers the question: How does the exploration budget affect performance? A detailed explanation of the results is provided in Section~\ref{sec:main-findings}.

\section{Case Study}
\label{app:case-study}

\begin{figure*}
    \centering
    \includegraphics[scale=0.43,angle =90]{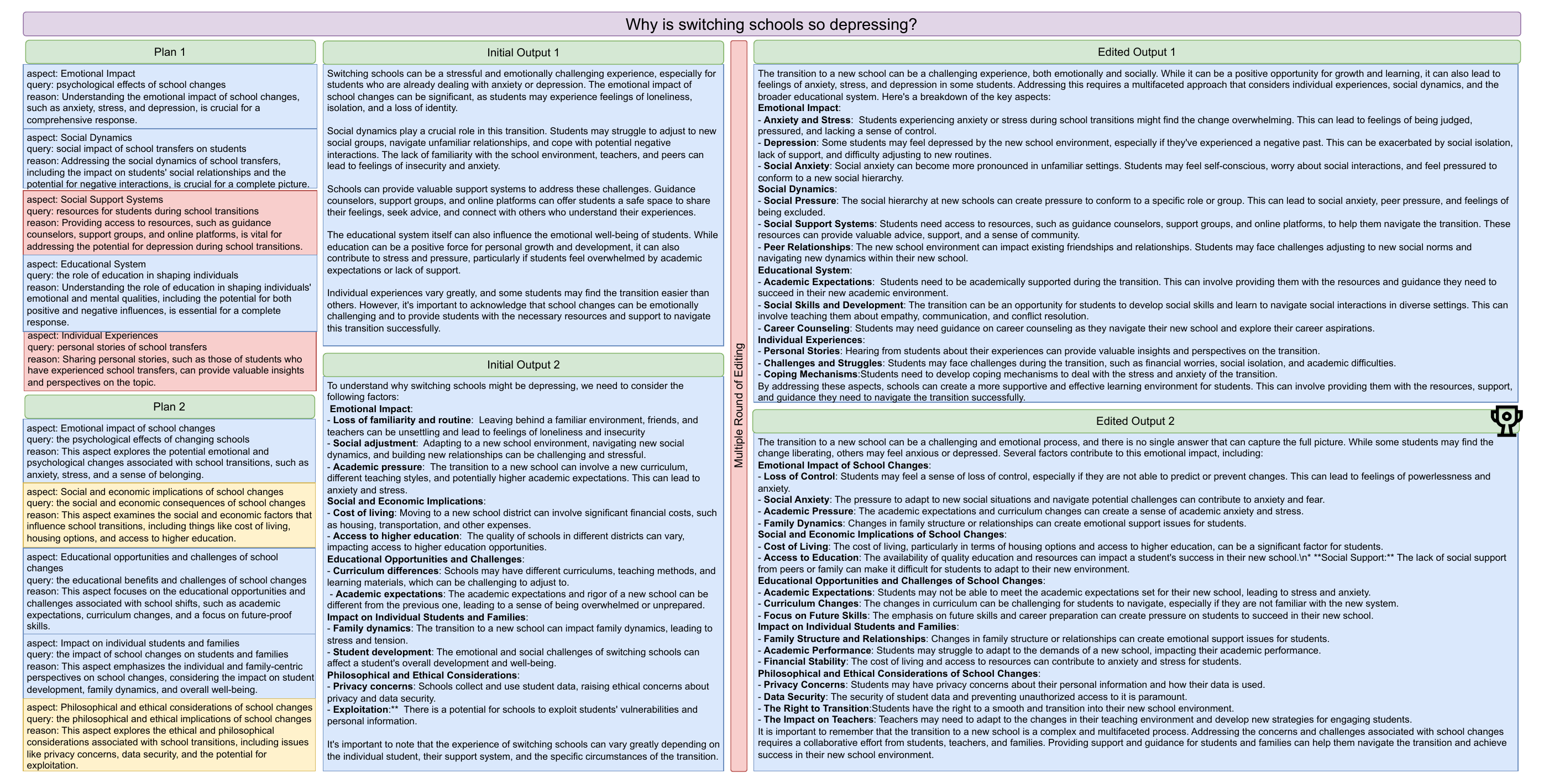}
    \vspace{-0.0cm}
    \caption{Case study on generated plans, responses, and edited responses by \ourmethod. The aspects that differ between the two plans are highlighted using different colors. The selected response is marked by a cup symbol.}
    \label{fig:case-study}
    \vspace{-0.4cm}
\end{figure*}

To provide a clearer understanding of how \ourmethod works, we present an output example for a query from the ANTIQUE dataset in Figure~\ref{fig:case-study}. Here, we generate two plans for global exploration, and for local exploitation, we iteratively edit the responses up to a maximum of 32 steps. As illustrated in Figure~\ref{fig:case-study}, the two plans share several steps in addressing the query while also considering unique aspects (unique aspects are highlighted in different colors, and shared steps are shown in the same color). For instance, second plan emphasizes the economic, philosophical, and ethical reasons behind depression following a school change, whereas the first plan focuses on mentioning individual experiences, examples, and social groups that can help alleviate such challenges. This difference in the plans resulted in two distinct initial responses in terms of both content and style. Next, the initial generated responses are refined by the editing model over multiple steps to produce the edited outputs depicted in Figure~\ref{fig:case-study}. An interesting observation is that the edited responses exhibit greater depth in categorizing various aspects and provide more detailed and structured explanations. This structuring is particularly noticeable in the first output. Initially, the first response was presented as paragraphs without utilizing markdown formatting or hierarchical organization for different aspects. However, the edited output introduces markdown elements and restructures the response, enhancing its coverage and factuality. Finally, the reward model selected the second edited output as the final response to the question. This choice reflects its superior coverage and factual accuracy, as evidenced by its ability to address a broader range of aspects while maintaining a high degree of factual correctness.

\end{document}